\documentclass[11pt,a4paper]{article}
\usepackage[hyperref]{eacl2021}
\usepackage{times}
\usepackage{latexsym}
\usepackage{booktabs}

\usepackage{microtype}
\usepackage{multicol}
\usepackage{multirow}
\usepackage{makecell}
\usepackage{graphicx}
\usepackage{subcaption}
\usepackage{todonotes}
\usepackage{amsmath,amsthm,amsfonts,amssymb}
\usepackage{enumitem}

\aclfinalcopy 

\newcommand{\SM}[1]{{\color{black}#1}} 

\title{Model Agnostic Answer Reranking System for \\Adversarial Question Answering}

\author{%
Sagnik Majumder \hspace{5mm} Chinmoy Samant \hspace{5mm} Greg Durrett\\
Department of Computer Science\\
The University of Texas at Austin\\
\texttt{\{sagnik, chinmoy, gdurrett\}@cs.utexas.edu}
}

\date{}

\begin{document}
\maketitle

\begin{abstract}
While numerous methods have been proposed as defenses against adversarial examples in question answering (QA), these techniques are often model specific, require retraining of the model, and give only marginal improvements in performance over vanilla models. In this work, we present a simple model-agnostic approach to this problem that can be applied directly to any QA model without any retraining. Our method employs an explicit answer candidate reranking mechanism that scores candidate answers on the basis of their content overlap with the question before making the final prediction. Combined with a strong base QA model, our method outperforms state-of-the-art defense techniques, calling into question how well these techniques are actually doing and strong these adversarial testbeds are.
\end{abstract}

\section{Introduction}
\begin{figure*}[!ht]
\centering 
\includegraphics[width=0.80\linewidth]{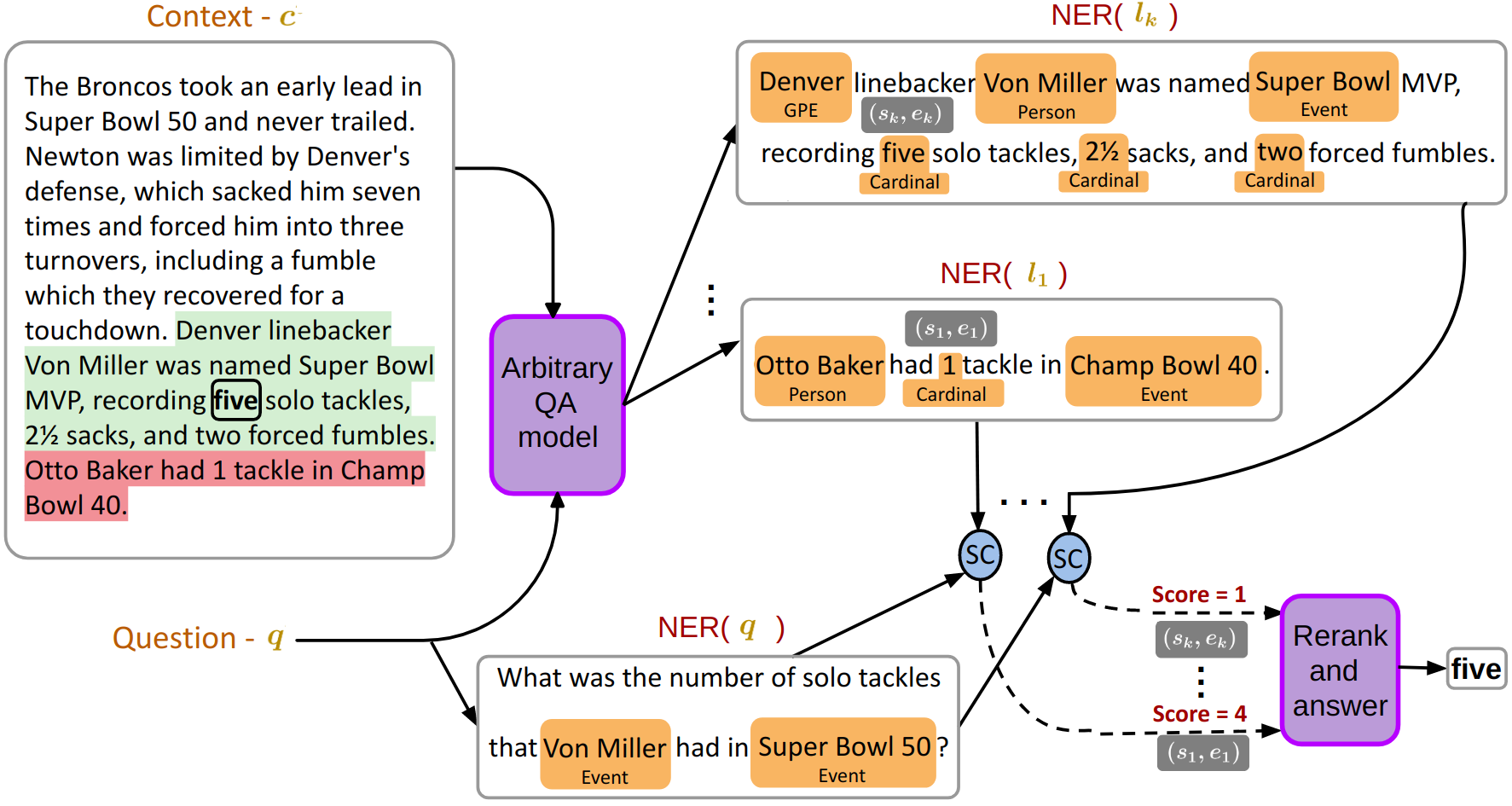}
\caption{Our model agnostic answer reranking system (MAARS). Given each answer option (right column), we extract named entities and compare them to named entities in the question. The overlap is used as a reranking feature to choose the final answer. The ground truth answer containing sentence is highlighted in green, the ground truth answer is boxed and the distractor sentence is highlighted in red.
}
\vspace*{-0.18in}
\label{fig:pipeline}
\end{figure*}
As reading comprehension datasets~\citep{richardson-etal-2013-mctest, Weston2015TowardsAQ, NIPS2015_5945, rajpurkar-etal-2016-squad, joshi-etal-2017-triviaqa} and models~\citep{NIPS2015_5846, DBLP:journals/corr/SeoKFH16, devlin-etal-2019-bert} have advanced, QA research has increasingly focused on out-of-distribution generalization~\citep{khashabi2020unifiedqa, talmor-berant-2019-multiqa}
and robustness. 
\citet{jia-liang-2017-adversarial} and~\citet{wallace-etal-2019-universal} show that appending unrelated distractors to contexts can easily confuse a deep QA model, calling into question the effectiveness of these models. Although these attacks do not necessarily reflect a real-world threat model, they serve as an additional testbed for generalization: models that perform better against such adversaries might be expected to generalize better in other ways, such as on contrastive examples~\citep{gardner2020evaluating}.

In this paper, we propose a simple method for adversarial QA that explicitly reranks candidate answers predicted by a QA model according to a notion of content overlap with the question.
Specifically, by identifying contexts where more named entities are shared with the question, we can extract answers that are more likely to be correct in adversarial conditions.

The impact of this is two-fold.
First, our proposed method is model agnostic in that it can be applied post-hoc to any QA model that predicts probabilities of answer spans, without any retraining. Second \SM{but most important}, we demonstrate that even this simple named entity based question-answer matching technique can be surprisingly useful. We show that our method outperforms state-of-the-art but more complex adversarial defenses with both BiDAF~\citep{DBLP:journals/corr/SeoKFH16} and BERT~\citep{devlin-etal-2019-bert} on two standard adversarial QA datasets~\citep{jia-liang-2017-adversarial, wallace-etal-2019-universal}. The fact that such a straightforward technique works well calls into question how reliable current datasets are for evaluating actual robustness of QA models.

\section{Related Work}
\SM{
Over the years, various methods have been proposed for robustness in adversarial QA, the most prominent ones} being adversarial training~\citep{wang-bansal-2018-robust, lee-etal-2019-domain, yang2019improving}, data augmentation~\citep{welbl2020undersensitivity} and posterior regularization~\citep{zhou2019robust}. \SM{Among these,} we compare our method only with techniques that train on clean SQuAD \cite{wu-etal-2019-improving,yeh-chen-2019-qainfomax} \SM{for fairness}. \citet{wu-etal-2019-improving} use a syntax-driven encoder to model the syntactic match between a question and an answer.~\citet{yeh-chen-2019-qainfomax} use a prior approach~\citep{hjelm2018learning} to maximize mutual information among contexts, questions, and answers to avoid overfitting to surface cues. In contrast, our technique is more closely related to retrieval-based methods for open-domain QA~\citep{chen-etal-2017-reading, yang-etal-2019-end-end-open} 
and multi-hop QA~\citep{welbl-etal-2018-constructing, de-cao-etal-2019-question}: we show that shallow matching can improve the reliability of deep models against adversaries in addition to these more complex settings.

\SM{Methods for (re)ranking of candidate passages/answers have often been explored in the context of information retrieval~\citep{10.1145/2766462.2767738}, content-based QA~\citep{kratzwald-etal-2019-rankqa} and open-domain QA~\citep{wang2018evidence, lee-etal-2018-ranking}. Similar to our approach, these methods also exploit some measure of coverage of the query by the candidate answers or their supporting passages to decide the ranks. However, the main motive behind ranking in such cases is usually to narrow down the area of interest within the text to look for the answer. On the contrary, we use a reranking mechanism that allows our QA model to ignore distractors in adversarial QA and can also provide model- and task-agnostic behavior unlike the commonly used learning-based (re)ranking mechanisms.

In yet another related line of research,~\citep{chen-etal-2016-thorough, kaushik-lipton-2018-much} reveal the simplistic nature and certain important  shortcomings of popular QA datasets. ~\citet{chen-etal-2016-thorough} conclude that the simple nature of the questions in the CNN/Daily Mail reading comprehension dataset~\citep{NIPS2015_afdec700} allows a QA model to perform well by extracting single-sentence relations. \citet{kaushik-lipton-2018-much} perform an extensive study with multiple well-known QA benchmarks to show several troubling trends: basic model ablations, such as making the input \textit{question-} or \textit{passage-}only, can beat the state-of-the-art performance, and the answers are often localized in the last few lines, even in very long passages, thus possibly allowing models to achieve very strong performance through learning trivial cues. Although we also question the efficacy of well-known adversarial QA datasets in this work, our core focus is on exposing certain issues specifically with the design of the adversarial distractors rather than the underlying datasets.  
}

\section{Approach}
\begin{table*}[!t]
\small\centering
\begin{tabular}{c|c|cc|cc}
\toprule
\multirow{2}{*}{\textbf{Model}} & \multirow{2}{*}{\textbf{Original}} & \multicolumn{2}{c|}{\textbf{AddSent}} & \multicolumn{2}{c}{\textbf{AddOneSent}}\\
& & \textbf{Adversarial} & \textbf{Mean} & \textbf{Adversarial} & \textbf{Mean}\\
\midrule
\makecell{BERT-S} & \textbf{89.4/82.1} & 40.9/35.9 & 68.0/61.7 & 54.6/48.4 & 74.1/67.2\\
\makecell{BERT-S + QAInfoMax} & 87.7/82.1 & 41.8/37.2 & 67.5/62.3 & 55.5/49.7 & 73.5/67.8\\
\makecell{BERT-S + MAARS} & 80.2/71.1 & \textbf{61.2/53.6} & \textbf{71.8/63.4} & \textbf{71.3/63.5} & \textbf{76.3/67.8}\\
\bottomrule
\end{tabular}
\caption{\label{table:bert_ours_vs_baselines} AddSent and AddOneSent results with BERT-S. MAARS outperforms the vanilla and baseline models on adversarial data but its performance drops a bit on the original data due to constrained reranking of answers.}
\vspace*{-0.18in}
\end{table*}
Neural QA models are usually trained in a supervised fashion on labeled examples of contexts, questions, and answers to predict answer spans; we represent these as $(s, e)$ tuples, where $s$ represents the sentence and $e$ the candidate span. Prior work
~\citep{lewis2018generative, mudrakarta-etal-2018-model, yeh-chen-2019-qainfomax, chen-durrett-2019-understanding} has noted that the end-to-end paradigm can overfit superficial biases in the data causing learning to stop when simple correlations are sufficient for the model to answer a question confidently. By explicitly enforcing content relevance between the predicted answer-containing sentence and the question, we can combat this poor generalization.

Specifically, we explicitly score the candidate sentences as per the word-level overlap in named entities common to both the question and a sentence. We refer to our method as Model Agnostic Answer Reranking System (MAARS).

Figure~\ref{fig:pipeline} illustrates the workflow of MAARS.
MAARS can be applied to any arbitrary QA model that predicts answer span probabilities.
First, we use the base QA model to compute the $n$ best answer spans $\mathcal{A} = \{(s_{1}, e_{1}),\ldots, (s_{n}, e_{n})\}$ for a context-question pair $(c, q)$ where $n$ is a hyperparameter. Any answer span not lying in a single sentence is broken into subspans that lie in separate sentences and $\mathcal{A}$ is updated accordingly.

Next, we extract the set of candidate sentences $\mathcal{L}$ from the context containing these $n$ answer spans. For the question and each sentence, we compute a set of named entity chunks using an open-source AllenNLP~\citep{Gardner2017AllenNLP} NER model.
We then compute the set of words inside named entity chunks from each candidate sentence $\text{NER}(l_{k}) \text{ }\forall \text{ }l_{k}\in \mathcal{L}$ and the question $\text{NER}(q)$; note that NER($\cdot$) refers to a set of words and not a set of named entities. Each candidate sentence $l_{k}$ is then given a score $\text{SC}(l_{k}) = \text{NER}(l_{k}) \cap \text{NER}(q)$ and the answer spans are reranked per the scores of the sentences containing them. In the case of ties or if there are multiple spans in the same candidate sentence, they are reranked among themselves according to the original ordering as per the QA model. Finally, the span with the highest rank after reranking is chosen as the final answer.

Compared to the base QA model, this approach only relies on an additional NER model that can be used without any retraining of the base model. Note that the architecture doesn't depend on any specific tagger, and the other content matching models like word matching could also be used in the system here.
\vspace*{-0.1in}

\section{Experiments}
\begin{table}[!tb]
\small
\centering
\begin{tabular}{c|c|cc}
\toprule
\multirow{2}{*}{\textbf{Model}} & \multirow{2}{*}{\textbf{Original}} & \multicolumn{2}{c}{\textbf{AddSent}}\\
& & \textbf{Adv.} & \textbf{Mean}\\\midrule
\makecell{BiDAF} & \makecell{72.4/62.4} & \makecell{21.4/16.0} & \makecell{49.9/42.0}\\
\makecell{BiDAF + SLN} & 72.3/62.4 & 22.8/17.2 & 50.5/42.5\\
\makecell{BiDAF + MAARS} & \makecell{\textbf{72.3/62.9}} & \makecell{\textbf{45.4/38.0}} & \makecell{\textbf{60.4/51.9}}\\
\bottomrule
\end{tabular}
\caption{\label{table:bidaf_ours_vs_baselines} AddSent results with BiDAF. Here, MAARS beats the vanilla and baseline models across all metrics.}
\vspace*{-0.18in}
\end{table}

\subsection{Evaluation settings}\label{subsec:exp_eval}
\paragraph{Datasets and baselines.} We evaluate MAARS on two well-known adversarial QA datasets built on top of SQuAD v1.1: Adversarial SQuAD~\citep{jia-liang-2017-adversarial} and Universal Adversarial Triggers~\citep{wallace-etal-2019-universal}. \SM{For brevity, we don't include the adversarial distraction generation process for either of the datasets and point the interested reader to the original papers for exact details}. For Adversarial SQuAD, we test MAARS with both BiDAF and BERT and compare against state-of-the-art baselines on adversary types used in the original papers.
To the best of our knowledge, there is no pre-existing literature that proposes a defense technique for Universal Triggers. We also find that it fails to degrade the performance of our vanilla BERT model, probably because the attacks were originally generated for BiDAF. Thus, we only evaluate on this dataset in the BiDAF setting, using all four triggers \emph{Who}, \emph{When}, \emph{Where} and \emph{Why}.

For BiDAF, we compare MAARS against the Syntactic Leveraging Network (SLN) by~\citet{wu-etal-2019-improving} on \emph{AddSent}. SLN encodes predicate-argument structures from the context and question, a conceptually similar structure matching approach as MAARS but trained end-to-end with many more parameters. For BERT,
we benchmark MAARS against QAInfoMax~\citep{yeh-chen-2019-qainfomax} on \emph{AddSent} and \emph{AddOneSent}. In addition to the standard loss for training QA models, QAInfoMax adds a loss to maximize the mutual information between the learned representations of words in context and their neighborhood, and also between those of the answer spans and the question. 
\vspace*{-0.2in}
\\\paragraph{Implementation details.}
We use the uncased base (single) pretrained BERT from HuggingFace~\citep{Wolf2019HuggingFacesTS} and finetune it using Adam with weight decay~\citep{loshchilov2018decoupled} optimizer and an initial learning rate of $3e^{-5}$ on SQuAD~\citep{rajpurkar-etal-2016-squad} v1.1 for 2 epochs for both vanilla BERT and BERT + QAInfoMax. We set the training batch size to 5 and the proportion of linear learning rate warmup for the optimizer to 10\%.

Our BiDAF~\citep{DBLP:journals/corr/SeoKFH16} model has a hidden state of size 100 and takes 100 dimensional GloVe~\citep{pennington-etal-2014-glove} embeddings as input. For character-level embedding, it uses 100 one-dimensional convolutional filters, each with a width of 5. A uniform dropout~\citep{JMLR:v15:srivastava14a} of 0.2 is applied at the CNN layer for character embedding, all LSTM~\citep{doi:10.1162/neco.1997.9.8.1735} layers and at the layer before the logits. We train it with AdaDelta~\citep{DBLP:journals/corr/abs-1212-5701} and an initial learning rate of 0.5 for 50 epochs. We set the training batch size to 128. For our Syntactic Leveraging Network, we follow the exact hyperparameter settings of~\citep{wu-etal-2019-improving}.

Other hyperparameters common to both BERT and BiDAF include an input sequence length of 400, maximum query length of 64, and 40 predicted answer spans per context-question pair. For NER tagging, we use an ELMo-based implementation from AllenNLP ~\citep{Gardner2017AllenNLP} that has been finetuned on CoNLL-2003 ~\citep{tjong-kim-sang-de-meulder-2003-introduction}.  Finally, we set the value of $n$ (the number of candidates considered for reranking) in MAARS to 10 across all our experiments.
\vspace*{0.2in}
\begin{table}[!tb]
\small
\centering
\begin{tabular}{c|c|c}
\toprule
\textbf{Adv. type} & \textbf{BiDAF} & \textbf{BiDAF + MAARS}\\\midrule
Who & 74.4/67.3 & \textbf{76.3/68.9} \\ 
When & 80.1/75.5 & \textbf{81.8/77.1} \\ 
Where & 63.5/52.8 & \textbf{68.8/56.7} \\
Why & \textbf{51.9/34.1} & 51.6/34.1 \\ 
\bottomrule
\end{tabular}
\caption{\label{table:triggers}Results on Universal Triggers with BiDAF (BERT-specific triggers unavailable publicly). MAARS is better than the vanilla model for most adversaries but with smaller performance gains than Adversarial SQuAD.}. 
\vspace*{-0.2in}
\end{table}
\vspace*{-0.3in}
\begin{figure*}[!tb]
\centering
\subcaptionbox{Wrong top candidate picked\label{subfig:fail_wrongTopSent2}}{\includegraphics[width=0.33\textwidth]{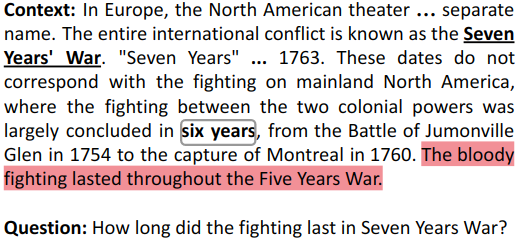}}%
\hfill
\subcaptionbox{Lack of attention to question type\label{subfig:fail_wrongAttend2}}{\includegraphics[width=0.33\textwidth]{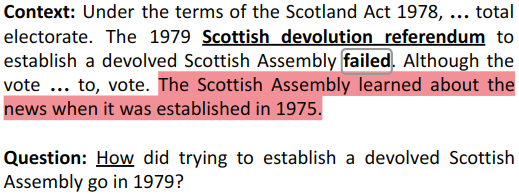}}%
\hfill
\subcaptionbox{Multiple similar spans co-occur \label{subfig:fail_mutliSpan2}}{\includegraphics[width=0.33\textwidth]{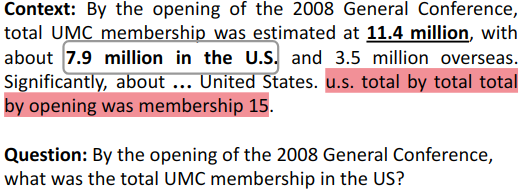}}%
\vspace{-0.1in}
\caption{Common failure cases for MAARS. The distractor sentence is highlighted in red, the predicted answer is underlined and the ground truth answer is boxed.}
\vspace*{-0.18in}
\label{fig:failures}
\end{figure*}

\subsection{Results}

In all our results tables, we report the macro-averaged F1 and exact match (EM) scores separated by a slash in each cell. In Tables~\ref{table:bert_ours_vs_baselines} and~\ref{table:bidaf_ours_vs_baselines}, \textbf{Original} and \textbf{Adversarial} (\textbf{Adv.}) refer to a model's performance on only clean and only adversarial data respectively. \textbf{Mean} denotes the weighted mean of the \textbf{Original} and \textbf{Adversarial} scores, weighted by the respective number of samples in the dataset. Both \emph{AddSent} and \emph{AddOneSent} have 1000 clean and 787 adversarial instances.
\vspace*{-0.05in}

\paragraph{Adversarial SQuAD.} Table~\ref{table:bert_ours_vs_baselines} shows the results with BERT-single (-S) on \emph{AddSent} and \emph{AddOneSent}. MAARS outperforms both the vanilla model and QAInfoMax on both \textbf{Adversarial} and \textbf{Mean} metrics. The performance gains are also substantial, especially on \textbf{Adversarial} where MAARS improves F1 over QAInfoMax by about 20 points on \emph{AddSent} and 16 points on \emph{AddOneSent}. This clearly shows that our method is much more capable of avoiding distractors in data and it is a much stronger defense technique in this setting. For both QAInfoMax and MAARS there is a drop in performance on clean data, but the drop for MAARS is larger. This drop naturally arises from the simplicity of the heuristic: matching words in named entities with the question sometimes assigns a higher score to a candidate sentence which has a higher overlap in terms of named entities with the question but doesn't contain the right answer.
One such example where MAARS fails to pick the correct top candidate after reranking is shown 
in fig.~\ref{subfig:fail_wrongTopSent2}.

Table \ref{table:bidaf_ours_vs_baselines} details the results with BiDAF on $AddSent$.\footnote{BiDAF + MAARS gives a similar F1/EM trend on \emph{AddOneSent} (\textbf{Adversarial}:  46.1/38.5, \textbf{Mean}: 60.8/52.1).}
Here, we also see significant performance gains over the vanilla model and the SLN baseline. MAARS results in an increase in adversarial F1 by 24 points over vanilla BiDAF and about 22 points over BiDAF + SLN. Interestingly, the performance on clean data doesn't drop as in the case of BERT. This difference may be a result of BiDAF using more surface word matching itself, leading to a closer alignment between its predictions and the reranker's choices. However, note that our simple heuristic still performs well even with a complex model like BERT.

\paragraph{Discussion.} Overall, our results on this dataset look promising for both BERT and BiDAF despite our method's inherent simplicity. This raises two questions. First, how effective is the Adversarial SQuAD dataset as a testbed for adversarial attacks? When a simple method can achieve large gains, we cannot be sure that more complex methods are truly working as advertised rather than learning such heuristics. Second, how effective are these current defenses? They underperform a simple heuristic in this setting; however, because the full breadth of possible adversarial settings has not been explored, it's hard to get a holistic sense of which methods are effective. Additional settings are needed to fully contrast these techniques.
\vspace*{-0.06in}

\paragraph{Universal Adversarial Triggers.} We create a  dataset that has purely adversarial instances using the open-source code from~\citet{wallace-etal-2019-universal} and present the results in Table~\ref{table:triggers}.
In particular, we append the following distractors for different adversary types. The target answers in the distractors have been bolded. 
\begin{itemize}[leftmargin=2ex]
\item \emph{Who}: \textcolor{red}{how ] ] there \textbf{donald trump} ; who who did}
\item \emph{When}: \textcolor{red}{; its time about \textbf{january 2014} when may did british}
\item \emph{Where}: \textcolor{red}{; : ’ where \textbf{new york} may area where they}
\item \emph{Why}: \textcolor{red}{why how ; known because : \textbf{to kill american people} .}
\end{itemize}
Due to unavailability of prior work on trigger-specific defense and BERT-specific triggers, we report only vanilla BiDAF and BiDAF with MAARS. F1 drops by a small amount (0.3 points) from BiDAF to BiDAF with MAARS while the EM score doesn't change at all for \emph{Why}. The scores improve by around 1-2 points for the other adversary types. However, the gains are much lower in comparison to Adversarial SQuAD. These results indicate the promise of simple defenses, but more exhaustive evaluation of defenses on different types of attacks is needed to draw a more complete picture of the methods' generalization abilities.

\subsection{Failure cases}
Besides the instances where the primary error source is picking a wrong top candidate (refer to Fig.~\ref{subfig:fail_wrongTopSent2}), we notice two other common failure case types with MAARS. One directly stems from MAARS' inability to attend to the question type during reranking. In Fig.~\ref{subfig:fail_wrongAttend2}, the question word is \emph{How} but MAARS picks \emph{Scottish devolution referendum} which is not the appropriate type of answer here.
The other type of failure occurs when multiple similar span types are present in the same candidate, thus creating ambiguity for the base QA model. In the example shown in Fig.~\ref{subfig:fail_mutliSpan2}, the QA model fails to distinguish between the two spans and retrieve specific information about \emph{the US}. 
Better base QA models may resolve these issues, or a more powerful reranker could also be used. However, rerankers learned end-to-end would suffer from the same issues as BERT and require additional engineering to avoid overfitting the training data.


\section{Conclusion}
In this work, we introduce a simple and model agnostic post-hoc technique for adversarial question answering (QA) that predicts the final answer after re-ranking candidate answers from a generic QA model as per their overlap in relevant content with the question.
Our results show the potential of our method through large performance gains over vanilla models and state-of-the-art methods. We also analyze common failure points in our method. {Finally, we reiterate that our \SM{main} contribution is not the heuristic defense itself but rather its ability to paint a more complete picture of the current state of affairs in adversarial QA. We seek to illustrate that our current adversaries are not strong and generic enough to attack a wide variety of QA methods, and we need a broader evaluation of our defenses to meaningfully gauge our progress in adversarial QA research.

\bibliography{anthology,eacl2021}
\bibliographystyle{acl_natbib}

\end{document}